\documentclass{article}

\PassOptionsToPackage{numbers, compress}{natbib}
\usepackage[dblblindworkshop, final]{neurips_2026}
\workshoptitle{VLM4RWD}

\usepackage[utf8]{inputenc} 
\usepackage[T1]{fontenc}    
\usepackage{url}            
\usepackage{booktabs}       
\usepackage{amsfonts}       
\usepackage{nicefrac}       
\usepackage{microtype}      
\usepackage{xcolor}         

\usepackage{graphicx}
\usepackage{todonotes}
\usepackage{booktabs}
\usepackage{amsmath}
\usepackage{subcaption}
\usepackage{hyperref}

\definecolor{linkblue}{HTML}{2980B9}

\hypersetup{
    colorlinks=true,
    linkcolor=black,
    citecolor=black,
    urlcolor=linkblue
}

\title{Foveated Compression: Selective High-Resolution Preservation for Token-Efficient VLMs}

\author{
    Donghyun Han$^{1*\dagger}$, \qquad
    Jangho Park$^{2*}$, \qquad 
    Yuseok Bae$^{1}$ \\
    $^{1}$ETRI, South Korea \\
    $^{2}$Kyung Hee University, South Korea \\
}

\begin{document}

\maketitle

\begingroup
\renewcommand{\thefootnote}{\fnsymbol{footnote}}
\footnotetext[1]{Equal contribution.}
\footnotetext[2]{Corresponding author.}
\endgroup

\begin{abstract}
Visual tokens are a major source of inference cost in vision-language models, yet simple image downsampling remains a surprisingly strong compression baseline. This raises a complementary question: under a fixed token budget, where should visual fidelity be preserved? We introduce Foveated Compression, which encodes a full-resolution image once and represents it with a mixture of native- and compressed-resolution visual tokens. A behaviorally self-distilled Foveated Merger compresses local visual tokens while preserving compatibility with their native counterparts, and a lightweight Foveated Selector chooses one of nine spatial cells to retain at native resolution using exhaustive budget-matched intervention supervision. Token budgets count visual tokens entering the language model, which govern prefill cost and KV-cache memory; the vision encoder still runs once at full resolution. At 11.11\% visual tokens, uniform Foveated Compression shows no significant paired difference from iso-token downsampling. At 20.99\%, the learned selector significantly outperforms random and fixed allocation, but remains below strong whole-image resizing, showing that localized fidelity is not universally preferable. Within the same 3×3, K=1 action space, a budget-matched region-choice oracle reaches 82.73 macro accuracy versus 69.61 for the learned selector, revealing substantial unrealized headroom. Matched probing further shows that signals predicting when compression breaks the answer are substantially more accessible after language-model computation than to the lightweight prefill-free selector. These results expose complementary bottlenecks in region selection and compressed-region fidelity. Our code and evaluation scripts are available on the project page: \url{https://github.com/Donghyun99/Foveated-Compression}
\end{abstract}


\section{Introduction}
\label{sec:intro}

Vision-language models (VLMs) increasingly rely on high-resolution visual inputs for fine-grained perception, document understanding, and visually grounded reasoning~\cite{qwen25vl,fastvlm2025}. Higher resolution, however, produces hundreds or thousands of visual tokens, often far more than the accompanying text, increasing language-model prefill cost and memory use~\cite{fastvlm2025,sparsevlm2025}. This has motivated visual-token pruning~\cite{dynamicvit2021,fastv2024,sparsevlm2025,visionthink2025,feather2025}, merging~\cite{tome2023,prumerge2025,visiontrim2026}, resampling~\cite{flamingo2022}, and learned compression~\cite{vocollama2025,focusllava2024}. The central challenge is not reducing the sequence, but preserving the visual detail needed by high-resolution tasks.

VTC-Bench~\cite{liao2026vtcbench} complicates this picture by showing that simple image downsampling can match or outperform more elaborate compression methods. Downsampling preserves global image coverage while reducing spatial fidelity uniformly, suggesting that native resolution is rarely needed everywhere. This motivates a different question: rather than only asking \emph{how many} visual tokens to retain~\cite{adaptvision2026}, can we decide \emph{where} a limited fidelity budget should be spent?

Our diagnostic experiments indicate substantial spatial variation in the value of high resolution. Many examples remain correct after aggressive resolution reduction, while others recover only when fine detail is restored. Exhaustive $K{=}1$ interventions over a $3\times3$ spatial grid show that the best region varies across examples, though many examples are insensitive to which region is preserved. When region choice does matter, a well-chosen single region can recover substantial performance lost under uniform compression. Pixel-space selective magnification~\cite{teva2025, vstar2024, zoomeye2025, qzoom2026} offers one route to such allocation, but may revisit content already represented by the global view, require additional high-resolution visual processing, and lose relevant evidence when an unhelpful region is selected and the remaining content is not represented. These observations motivate allocating fidelity directly within a single encoded visual representation.

\begin{figure}[t]
\centerline{\includegraphics[width=0.9\linewidth]{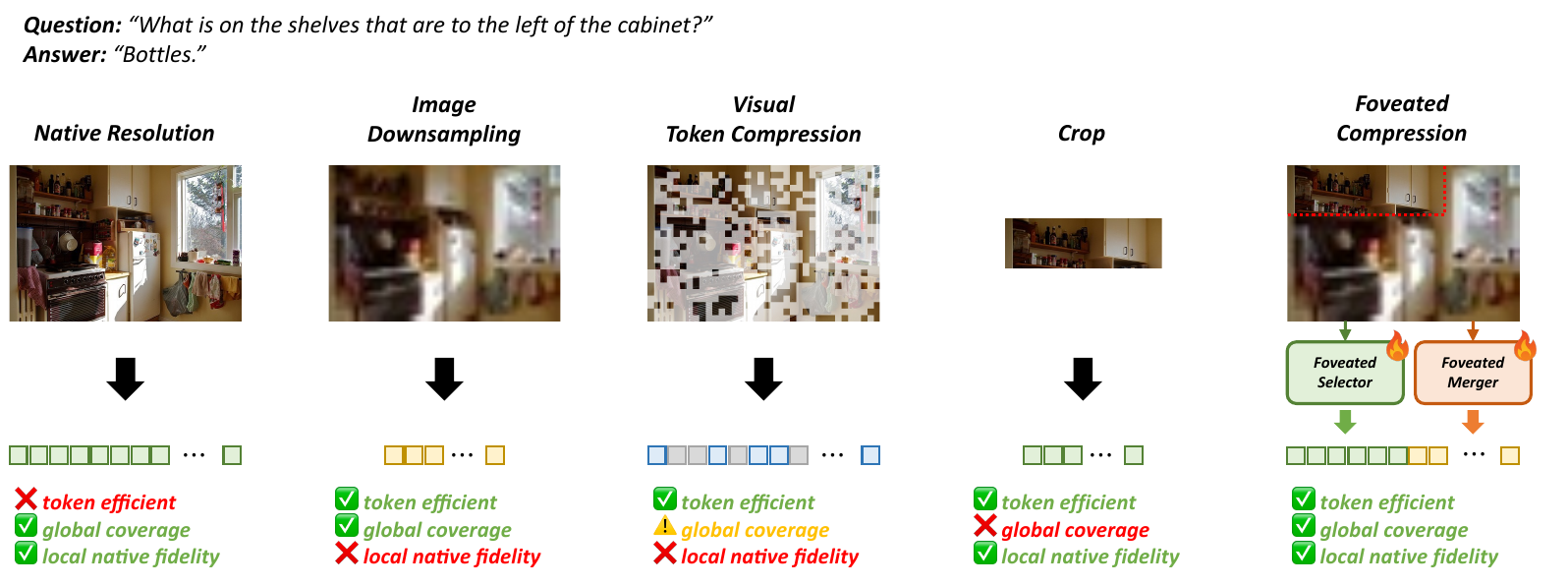}}
\caption{Foveated Compression preserves one selected region at native visual fidelity while keeping the rest of the image compressed within a single full-resolution encoding. Unlike whole-image downsampling or selective cropping, it maintains global coverage and enables native-token restoration without additional high-resolution visual encoding.}
\vspace{-0.5cm}
\label{fig:teaser}
\end{figure}

We introduce \emph{Foveated Compression}, which performs spatial fidelity allocation directly in feature space (Figure~\ref{fig:teaser}). A Foveated Merger compresses each $3\times3$ block of native visual tokens into one token and is trained by behavioral self-distillation from the frozen model's full-resolution predictions. Unlike image downsampling, this feature-space representation retains direct access to the native tokens from the same full-resolution encoding, allowing selected compressed regions to be replaced with their native counterparts without re-encoding the image. Uniform Foveated Compression uses only 11.11\% of the native visual-token count while retaining compressed coverage of the entire image. Throughout, token efficiency refers to language-model-side visual tokens: Foveated Compression still runs the vision encoder once at full resolution, so its savings apply to prefill and KV-cache memory rather than to encoder cost.

A lightweight Foveated Selector ranks nine equal spatial cells and preserves the highest-scoring cell at native token resolution under a fixed $K{=}1$ budget. Supervision is obtained by exhaustively evaluating all nine budget-matched interventions and training the selector on the resulting winner sets. The selected cell replaces its compressed counterpart while the other eight remain compressed, yielding an exact 20.99\% visual-token budget regardless of which region is chosen. The selector requires neither an additional language-model forward pass nor an additional high-resolution vision-encoder pass, and unselected regions remain represented in compressed form.

Our results reveal both the promise and the current limits of spatial fidelity allocation. At 11.11\% visual tokens, uniform Foveated Compression matches iso-token downsampling within statistical uncertainty (95\% CI [-0.586, 0.521]). At 20.99\%, the learned selector significantly outperforms random and training-only fixed-region allocation, showing that useful spatial preferences can be learned under a fixed budget. However, strong whole-image resizing remains better on average, indicating that localized native preservation is not universally preferable to uniform resolution reduction. Within the same $3\times3$, $K{=}1$ action space, a budget-matched region-choice oracle reaches 82.73 macro accuracy compared with 69.61 for the learned selector, exposing substantial selection headroom. Matched probing further shows that signals predicting when compression breaks the answer are substantially more accessible after language-model computation than to the lightweight prefill-free selector. Together, these results point to complementary challenges in region selection and compressed-region fidelity.

Our contributions are:
\begin{itemize}

\item We introduce Foveated Compression, a mixed-resolution feature-space representation that combines aggressive global compression with native-resolution preservation in one selected region, while requiring only a single full-resolution visual encoding.

\item We introduce a behaviorally self-distilled Foveated Merger that compresses $3\times3$ native visual-token blocks into one token while preserving direct compatibility with native tokens, enabling mixed-resolution replacement within the same encoded image.

\item We formulate fixed-budget spatial fidelity allocation using exhaustive, budget-matched $K{=}1$ interventions and tie-aware supervision, enabling controlled comparisons among learned, random, fixed, and oracle region allocation under identical token cost.

\item We provide a diagnostic analysis of spatial fidelity allocation: a budget-matched oracle reveals large unrealized headroom, and matched probing shows that signals predicting compression failure are weakly accessible before language-model computation but strongly accessible after it, identifying where current prefill-free selection is limited.

\end{itemize}


\section{Related Work}
\label{sec}

\subsection{Visual Token Compression}

Visual-token reduction methods include pruning~\cite{dynamicvit2021,fastv2024,sparsevlm2025,promprune2026,visionthink2025,feather2025}, merging~\cite{tome2023,prumerge2025,libramerging2025,visiontrim2026}, pooling or resampling~\cite{flamingo2022}, and learned or adaptive compression~\cite{vocollama2025,focusllava2024,dymu2025,adaptvision2026}. VoCo-LLaMA~\cite{vocollama2025} distills language-model attention into learned compression tokens, FocusLLaVA~\cite{focusllava2024} assigns adaptive local visual scales before instruction-guided token selection, and DyMU~\cite{dymu2025} adapts token length to image content.

Recent work has also questioned whether increasingly elaborate compression consistently improves visual-token efficiency. VTC-Bench~\cite{liao2026vtcbench} shows that simple image downsampling can match or outperform learned compression methods, while benchmarks such as MMStar~\cite{mmstar} emphasize genuine visual dependence. We extend this perspective from global token reduction to spatial fidelity allocation: rather than varying only the number of retained tokens, we use exhaustive, budget-matched interventions to study whether performance depends on where native visual fidelity is preserved.

Matryoshka Multimodal Models ~\cite{m3_2025} learn nested coarse-to-fine visual tokens from a single encoding, allowing the granularity to be chosen at inference time, and report a substantial gap between an oracle granularity choice and fixed-granularity baselines. Our work is complementary along a different axis: M3 selects how many tokens represent the whole image at a uniform granularity and trains the language model, whereas we keep the backbone frozen and select where native fidelity is preserved within a single mixed-resolution sequence. Our region-choice oracle therefore measures spatial-allocation headroom at a fixed token count, rather than granularity-selection headroom.

Foveated Compression differs in its representation, supervision, and evaluation setting. We keep the vision-language backbone frozen, in contrast to end-to-end instruction tuning approaches such as FocusLLaVA~\cite{focusllava2024}, and train the Foveated Merger to match the frozen full-resolution model's output distribution. The Foveated Selector is supervised directly by exhaustive region interventions, while a fixed-$K{=}1$ design holds visual-token cost constant across candidate regions and enables controlled comparisons among learned, random, fixed, and oracle allocations. Unlike image downsampling, feature-space compression also preserves direct access to native tokens from the same full-resolution encoding, allowing selected regions to be restored at native fidelity without re-encoding the image. Attention-based pruning methods such as FastV~\cite{fastv2024} and SparseVLM~\cite{sparsevlm2025} select tokens using language-model signals and are complementary to our setting; combining them with mixed-resolution representations is a natural extension.

\subsection{Selective High-Resolution Processing}

A complementary line of work allocates visual fidelity non-uniformly by selectively revisiting or magnifying image regions~\cite{qzoom2026,visionopd2026,zwz2026,ergo2026,pixelreasoner2025}. V*~\cite{vstar2024} performs iterative visual search to locate task-relevant evidence, while ZoomEye~\cite{zoomeye2025} explores progressively magnified regions through hierarchical search. Dynamic tiling methods combine higher-resolution local crops with a coarse global view~\cite{llavanext2024}, and TEVA~\cite{teva2025} proposes relevant regions before applying dynamic patch sampling at multiple scales.

These approaches demonstrate the value of selectively allocating high-resolution processing, but may require additional region proposal, repeated visual processing, or renewed access to image pixels. Foveated Reasoning~\cite{foveatedreasoning2026} makes focusing an explicit action during decoding by re-encoding selected high-resolution evidence and injecting it into the reasoning trajectory. Foveated Compression instead performs allocation before decoding over a single full-resolution visual encoding: selected native tokens replace their compressed counterparts within the same sequence, while all unselected regions remain represented in compressed form.


\section{Proposed Method}
\label{sec:method}

\subsection{Overview}
\label{sec:overview}

Foveated Compression represents a full-resolution image using a mixture of native- and compressed-resolution visual tokens under a fixed budget. The frozen vision encoder is run once to obtain a native visual-token grid. A Foveated Merger reduces each local $3\times3$ block of native visual tokens to one compressed token, yielding 11.11\% of the native sequence. Separately, the image is partitioned into a coarse $3\times3$ grid of nine spatial cells, and a Foveated Selector chooses exactly one cell to preserve at native resolution. Replacing the compressed units in that cell with their corresponding native-token blocks yields an exact $K{=}1$ budget of 20.99\%. We choose the $3\times3$ grid and $K=1$ so that compression blocks and selection cells align exactly (Appendix~\ref{app:geometry}), making every candidate action cost the same number of tokens. This isolates the effect of region choice from token count, at the price of a coarse, fixed action space.

\begin{figure*}[t]
\centering
\includegraphics[width=\textwidth]{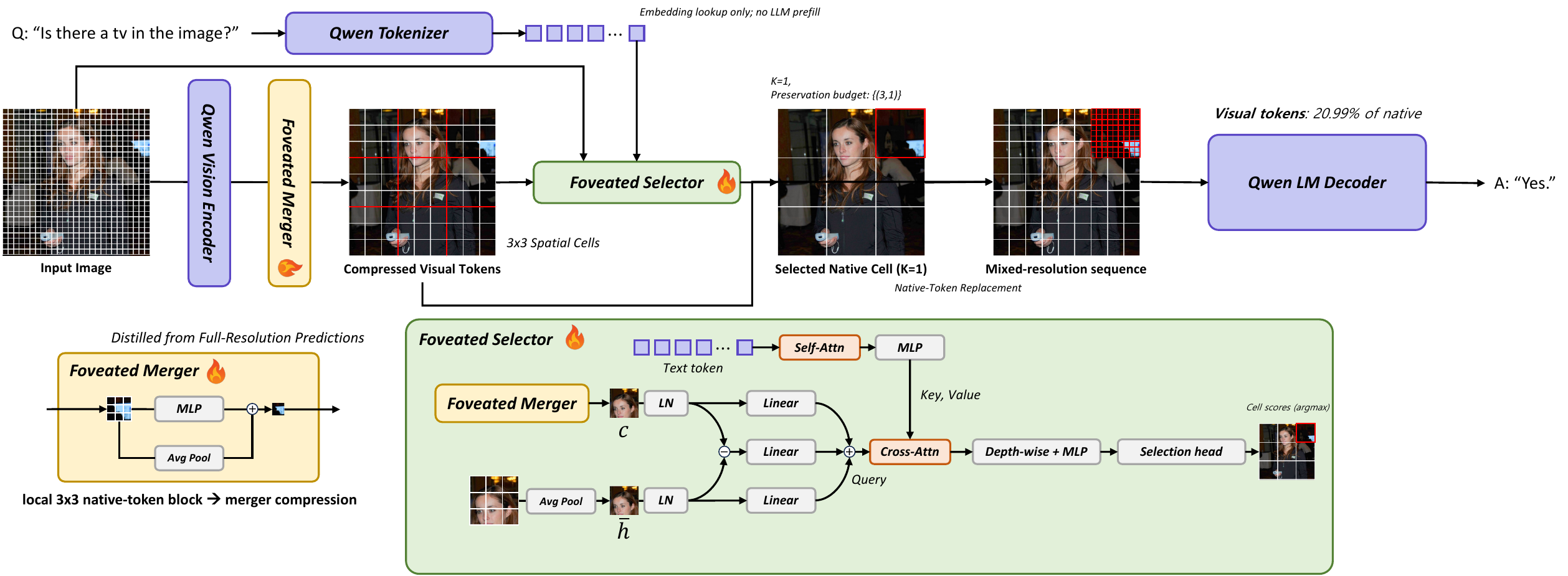}
\caption{Overview of Foveated Compression. The image is encoded once by the frozen vision backbone, the Foveated Merger produces a globally compressed representation, and the Foveated Selector restores one selected spatial cell with native visual tokens through in-place replacement.}
\label{fig:overview}
\end{figure*}

The vision-language backbone remains frozen throughout training. We first train and freeze the merger, then exhaustively evaluate the nine budget-matched $K{=}1$ interventions to supervise the selector. Figure~\ref{fig:overview} summarizes the pipeline.

\subsection{Foveated Merger}
\label{sec:merger}

The Foveated Merger operates on the vision-encoder patch grid before Qwen's frozen native visual merger. The encoder produces patch features $\mathbf{x}_i\in\mathbb{R}^{C}$ with $C{=}1280$. For each $3\times3$ neighborhood, we combine mean pooling with a learned residual transformation:
\begin{equation}
\tilde{\mathbf{x}}
=
\frac{1}{9}\sum_{i=1}^{9}\mathbf{x}_i
+
W_{\mathrm{head}}
\phi\!\left(
W_{\mathrm{fc}}
[
\mathrm{RMSNorm}(\mathbf{x}_1);
\ldots;
\mathrm{RMSNorm}(\mathbf{x}_9)
]
+
\mathbf{b}_{\mathrm{fc}}
\right)
+
\mathbf{b}_{\mathrm{head}},
\label{eq:foveated_merger}
\end{equation}
where $\phi$ denotes GELU,
$W_{\mathrm{fc}}:\mathbb{R}^{9C}\rightarrow\mathbb{R}^{2560}$,
and $W_{\mathrm{head}}:\mathbb{R}^{2560}\rightarrow\mathbb{R}^{C}$.
The resulting features pass through Qwen's frozen $2\times2$ visual merger to produce 3584-dimensional compressed tokens. Each compressed token corresponds spatially to a $3\times3$ block of native visual tokens and shares the same downstream interface. We randomly initialize $W_{\mathrm{fc}}$ and zero-initialize the residual head, so the module starts exactly from mean pooling. The merger has 32.77M trainable parameters.

\paragraph{Behavioral self-distillation.}
Only the merger is trained; the vision encoder, native visual merger, and language model remain frozen. The native model generates a greedy teacher answer $y_{1:L}$, and the compressed model is teacher-forced on the same sequence. Let $p_T^{(\ell)}$ and $p_S^{(\ell)}$ denote their next-token distributions at answer position $\ell$. We minimize
\begin{equation}
\mathcal{L}_{\mathrm{merge}}
=
\frac{1}{L}
\sum_{\ell=1}^{L}
D_{\mathrm{KL}}
\left(
p_T^{(\ell)}
\,\Vert\,
p_S^{(\ell)}
\right),
\label{eq:merger_distillation}
\end{equation}
thereby distilling the frozen model's full-resolution predictive behavior without ground-truth answer supervision.

\subsection{Foveated Selector}
\label{sec:selector}

The Foveated Selector estimates which of the nine coarse cells benefits most from native-resolution preservation. Each compressed unit $u$ corresponds to nine native visual tokens. Let $\mathbf{c}_u\in\mathbb{R}^{3584}$ denote the compressed token, $\mathcal{N}(u)$ its native-token block, and
\[
\bar{\mathbf{h}}_u
=
\frac{1}{9}
\sum_{i\in\mathcal{N}(u)}
\mathbf{h}_i
\]
their mean. Each unit also has a normalized 2D coordinate $\mathbf{p}_u\in\mathbb{R}^{2}$.

We construct a 512-dimensional visual descriptor
\begin{equation}
\mathbf{v}_u
=
P_n\!\left(\mathrm{LN}(\bar{\mathbf{h}}_u)\right)
+
P_c\!\left(\mathrm{LN}(\mathbf{c}_u)\right)
+
P_d\!\left(
\mathrm{LN}(\bar{\mathbf{h}}_u)
-
\mathrm{LN}(\mathbf{c}_u)
\right)
+
P_{xy}(\mathbf{p}_u),
\label{eq:selector_unit_feature}
\end{equation}
where $\mathrm{LN}$ denotes LayerNorm,
$P_n,P_c,P_d:\mathbb{R}^{3584}\rightarrow\mathbb{R}^{512}$
are learned projections, and
$P_{xy}:\mathbb{R}^{2}\rightarrow\mathbb{R}^{512}$
projects spatial position. The four terms encode native content, compressed content, compression-induced discrepancy, and location.

Question tokens are obtained from the frozen language-model embedding table, truncated to 48 tokens, projected to 512 dimensions, and encoded by a lightweight self-attention/FFN block. Each visual descriptor then attends to the question through eight-head cross-attention, with visual units as queries. A depthwise $3\times3$ convolution adds local spatial context, followed by a scoring MLP that predicts one logit $\ell_u$ per unit. The selector has 10.8M trainable parameters and requires no additional language-model forward pass. Although the architecture permits question-dependent scoring, Section~\ref{sec:ablations} finds no measurable gain attributable to question content alone.

Let $\mathcal{U}_c$ denote the compressed units contained in coarse cell $c$. We score each cell by
\[
s_c
=
\frac{1}{|\mathcal{U}_c|}
\sum_{u\in\mathcal{U}_c}\ell_u,
\qquad
c=1,\ldots,9,
\]
and select the highest-scoring cell. Whole-cell selection matches the intervention granularity and keeps all candidate actions at identical token cost.

\paragraph{Intervention-based supervision.}
For each training example, we evaluate all nine deterministic $K{=}1$ actions using the same generation and scoring protocol as evaluation. If $r_c$ is the score obtained by preserving cell $c$, the winner set is
$\mathcal{W}=\{c:r_c=\max_j r_j\}$.
All-nine ties are discarded because they provide no spatial preference, while partial ties are retained rather than collapsed to an arbitrary one-hot target. We minimize
\begin{equation}
\mathcal{L}_{\mathrm{selector}}
=
-\log
\frac{
\sum_{c\in\mathcal{W}}\exp(s_c)
}{
\sum_{j=1}^{9}\exp(s_j)
}.
\label{eq:tied_winner_loss}
\end{equation}
The merger and vision-language backbone remain frozen during selector training.

\subsection{Mixed-Resolution Construction}
\label{sec:mixed_resolution}

After selection, each compressed unit inside the chosen cell is replaced by its corresponding $3\times3$ block of native visual tokens; all other units remain compressed. Because both representations share the same 3584-dimensional interface, replacement requires neither re-encoding nor an additional projection. Unlike pixel-space downsampling, feature-space compression retains direct access to native tokens from the same full-resolution encoding.

If the native sequence contains $N$ visual tokens, restoring one of nine equal cells gives
\begin{equation}
N_{\mathrm{mix}}
=
\frac{N}{9}
+
\frac{8N}{81}
=
\frac{17N}{81},
\label{eq:mixed_token_count}
\end{equation}
or exactly 20.99\% of the native token budget.

Native and compressed tokens share the native visual-grid coordinate system for M-RoPE~\cite{qwen2vl}. Native tokens use unit-stride coordinates, while compressed tokens use the top-left coordinate of their corresponding $3\times3$ native block. Retaining absolute native-grid coordinates keeps other visual and subsequent text positions invariant across region choices.


\section{Experiments}
\label{sec:experiments}

\subsection{Experimental Setup}
\label{sec:setup}

We use Qwen2.5-VL-7B-Instruct as a frozen backbone. The merger is trained on GQA~\cite{gqa}, TextVQA~\cite{textvqa}, DocVQA~\cite{docvqa}, InfoVQA~\cite{infovqa}, ChartQA~\cite{chartqa}, and ScienceQA~\cite{scienceqa}, then frozen before generating exhaustive $K{=}1$ selector labels. We evaluate on GQA, MMBench-en/cn~\cite{mmbench}, MME~\cite{mme}, POPE~\cite{pope}, MMStar~\cite{mmstar}, OCRBench~\cite{ocrbench}, and ChartQA using official metrics. All conditions use geometry-matched preprocessing with image dimensions divisible by 252 pixels.

Uniform Foveated Compression uses 11.11\% of native visual tokens, while $K{=}1$ uses exactly 20.99\%. We compare against iso-token and matched-budget bicubic downsampling, random and training-only fixed-cell allocation, and $2\times$ whole-image downsampling. Native resolution and a test-time region-choice oracle serve as references. Matched-budget downsampling uses 19.82\% tokens in practice due to spatial alignment. All conditions use the same short-answer prompt, greedy decoding, and at most 16 generated tokens. We report the unweighted benchmark macro average; further details are provided in Appendix~\ref{app:implementation}.

\subsection{When and Where Does High Resolution Matter?}
\label{sec:hr_matter}

\begin{figure*}[t]
\centering
\begin{subfigure}[t]{0.34\textwidth}
    \centering
    \includegraphics[width=\linewidth]{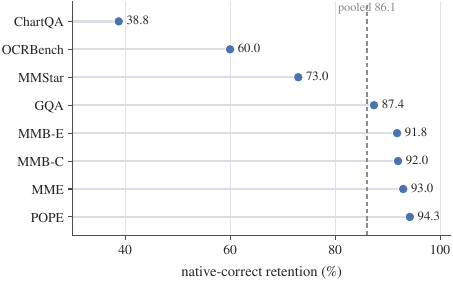}
    \caption{}
    \label{fig:analysis_retention}
\end{subfigure}
\hfill
\begin{subfigure}[t]{0.30\textwidth}
    \centering
    \includegraphics[width=\linewidth]{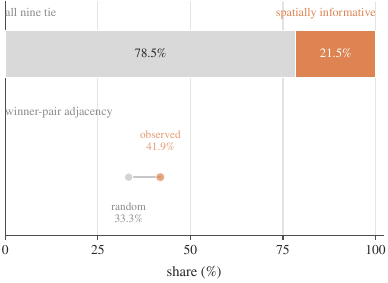}
    \caption{}
    \label{fig:analysis_spatial}
\end{subfigure}
\hfill
\begin{subfigure}[t]{0.30\textwidth}
    \centering
    \includegraphics[width=\linewidth]{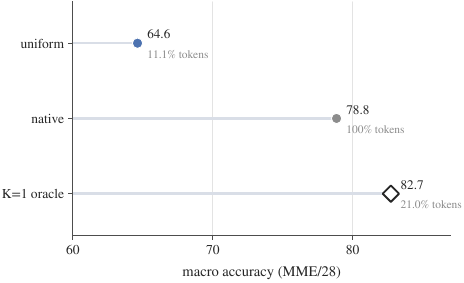}
    \caption{}
    \label{fig:analysis_headroom}
\end{subfigure}
\caption{High-resolution diagnostics. (a) Native-correct retention under uniform compression. (b) Spatial informativeness and winner-region adjacency. (c) Performance headroom from the budget-matched $K{=}1$ region-choice oracle.}
\label{fig:analysis}
\end{figure*}

Across 37,610 evaluation examples, uniform $9{:}1$ Foveated Compression retains 86.1\% of predictions that are correct at native resolution despite using only 11.11\% of the visual tokens. This \emph{native-correct retention} ranges from 38.8\% on ChartQA to 94.3\% on POPE (Figure~\ref{fig:analysis_retention}), showing substantial task-dependent sensitivity to compression.

Exhaustive $K{=}1$ interventions show that all nine cells receive the same task score on 78.5\% of examples, while the remaining 21.5\% are \emph{spatially informative} (Figure~\ref{fig:analysis_spatial}). Among informative cases, 41.9\% of winner pairs are horizontally or vertically adjacent, compared with $12/\binom{9}{2}=1/3$ for a random pair, suggesting non-random spatial structure.

A budget-matched region-choice oracle reaches 82.73 macro accuracy at 20.99\% tokens, versus 64.62 for uniform compression and 78.85 for the native reference (Figure~\ref{fig:analysis_headroom}). Its 3.87-point advantage over native (95\% CI $[3.388,4.366]$) reflects a test-time maximum over nine interventions and is interpreted only as an upper bound. The fixed $K{=}1$ action space nevertheless contains substantial spatial-allocation headroom.

\subsection{Foveated Compression Results}
\label{sec:main_results}

Table~\ref{tab:foveated_main_results} reports the main results on the same 37,610 examples.

\begin{table*}[t]
\centering
\small
\setlength{\tabcolsep}{4.2pt}
\caption{Main results on 37,610 examples across eight benchmarks. Bold and underlined values denote the best and second-best methods within the approximately 20\% fixed-budget comparison group. MME is divided by 28 only for the macro average.}
\resizebox{\textwidth}{!}{
\begin{tabular}{lcccccccccc}
\toprule
Method & Tokens & GQA & MMB-E & MMB-C & MME & POPE & MMStar & OCR & ChartQA & Macro \\
\midrule
Native & 100.00\% & 60.97 & 86.86 & 85.61 & 2356.7 & 87.74 & 60.93 & 80.20 & 84.28 & 78.85 \\
\midrule
Uniform Foveated Compression & 11.11\% & 58.28 & 82.51 & 82.01 & 2130.5 & 84.34 & 50.20 & 49.30 & 34.24 & 64.62 \\
Iso-Token Downsample & 11.11\% & 55.88 & 82.77 & 81.75 & 2240.8 & 82.69 & 51.07 & 57.30 & 27.44 & 64.86 \\
\midrule
Random $K{=}1$ & 20.99\% & \underline{58.40} & 84.18 & \underline{82.86} & \underline{2190.7} & 84.34 & \underline{52.33} & 55.10 & 48.36 & 67.98 \\
Fixed $K{=}1$ & 20.99\% & \textbf{58.90} & 83.69 & 82.70 & 2170.3 & 84.54 & 51.87 & 53.70 & \underline{54.92} & 68.48 \\
Foveated Selector & 20.99\% & 58.24 & \textbf{84.59} & 82.67 & 2186.7 & \textbf{84.92} & 52.13 & \underline{58.40} & \textbf{57.84} & \underline{69.61} \\
Matched-Budget Downsample & 19.82\% & 58.05 & \underline{84.45} & \textbf{83.34} & \textbf{2303.0} & \underline{84.77} & \textbf{54.53} & \textbf{67.00} & 53.48 & \textbf{70.99} \\
\midrule
$2\times$ Downsample & 24.57\% & 58.82 & 85.12 & 84.15 & 2298.7 & 84.90 & 56.00 & 70.50 & 57.84 & 72.43 \\
\midrule
$K{=}1$ Region-Choice Oracle & 20.99\% & 70.42 & 92.22 & 90.85 & 2500.8 & 91.82 & 70.67 & 77.60 & 78.92 & 82.73 \\
\bottomrule
\end{tabular}
}
\label{tab:foveated_main_results}
\end{table*}

At 11.11\% visual tokens, Uniform Foveated Compression and Iso-Token Downsample achieve nearly identical macro scores, 64.62 and 64.86, with no significant paired difference (95\% CI $[-0.586,0.521]$). Their task-level behavior differs: Foveated Compression is stronger on GQA, POPE, and ChartQA, whereas resizing is notably stronger on OCRBench and MME. The merger is therefore competitive with resizing on average while retaining the native-token compatibility required for mixed-resolution replacement.

At the fixed $K{=}1$ budget, the selector reaches 69.61, significantly outperforming random allocation at 67.98 by 1.62 points (95\% CI $[1.085,2.161]$) and training-only fixed allocation at 68.48 by 1.13 points (95\% CI $[0.610,1.632]$). Random allocation is nontrivial because informative selector-training labels contain 3.66 winning cells on average, corresponding to a 40.65\% random winner-membership rate. These results provide direct evidence that useful spatial preferences are learnable.

Strong whole-image resizing nevertheless remains better overall. Matched-Budget Downsample reaches 70.99 at 19.82\% tokens, outperforming the selector by 1.38 points (95\% CI $[0.707,2.062]$), while $2\times$ Downsample reaches 72.43 at 24.57\%, a 2.82-point advantage (95\% CI $[2.164,3.438]$). The comparison is task dependent: the selector exceeds matched-budget downsampling on ChartQA (57.84 vs.\ 53.48), whereas resizing has a large advantage on OCRBench (67.00 vs.\ 58.40) and MME. Thus, localized fidelity can help when relevant evidence is concentrated, but is not universally preferable.

The non-deployable $K{=}1$ oracle reaches 82.73 at the same 20.99\% budget, exceeding the learned selector by 13.12 points (95\% CI $[12.566,13.691]$). Although it selects using observed test outcomes, the large gap shows that substantially better region choices exist within the same action space. Full paired comparisons are provided in Appendix~\ref{app:additional_results}.

\subsection{Fidelity Allocation Analysis}
\label{sec:fidelity_analysis}

The oracle gap indicates substantial remaining selection headroom. To examine what information is accessible before language-model computation, we construct a matched binary task on 3,416 examples, with positives defined as examples where native inference is correct but uniform compression is incorrect. This predicts \emph{whether} compression fails rather than \emph{which} region should be selected, and therefore is not a direct measure of nine-way selection quality.

The raw prefill-free selector signal achieves an AUC of 0.5239 (95\% CI $[0.4796,0.5700]$), compared with 0.7651 (95\% CI $[0.7347,0.7937]$) for first-token decoder entropy on the same examples and target. A same-fold five-fold out-of-fold comparison yields 0.5476 (95\% CI $[0.5012,0.5929]$) versus 0.8072 (95\% CI $[0.7738,0.8391]$). Compression-sensitive information is therefore weakly accessible before prefill but substantially more predictable after language-model computation. Decoder-side signals arise only after generation and cannot be directly used by our single-pass selector. Further details are provided in Appendix~\ref{app:matched_auc}.

Region prediction is not the only limitation. On OCRBench, Foveated Compression trails downsampling at both 11.11\% tokens (49.30 vs.\ 57.30) and approximately 20\% tokens (58.40 vs.\ 67.00), whereas the selector exceeds Matched-Budget Downsample on ChartQA (57.84 vs.\ 53.48). These contrasts expose complementary bottlenecks in region prediction and compressed-region fidelity. Additional qualitative examples are provided in Appendix~\ref{app:qualitative}.

\subsection{Ablations}
\label{sec:ablations}

Table~\ref{tab:ablations} evaluates the main design choices using matched runs for controlled comparisons.

\begin{table*}[t]
\centering
\caption{Per-benchmark ablations. The merger block uses the 11.11\% uniform setting; selector ablations use a matched full-selector rerun (69.77 macro), distinct from the 69.61 main checkpoint; the two runs differ only in random seed. MME is divided by 28 only for the macro average. The Tie-Aware Supervision and Full Selector rows report the same run, listed once in each ablation block.}
\resizebox{\textwidth}{!}{
\begin{tabular}{lcccccccccc}
\toprule
Variant & Tokens & GQA & MMB-E & MMB-C & MME & POPE & MMStar & OCR & ChartQA & Macro \\
\midrule
Mean-Pooling Compression & 11.11\% & 51.76 & 79.65 & 78.33 & 2013.4 & 80.47 & 44.73 & 41.40 & 24.12 & 59.05 \\
Behaviorally Trained Merger & 11.11\% & \textbf{58.28} & \textbf{82.51} & \textbf{82.01} & \textbf{2130.5} & \textbf{84.34} & \textbf{50.20} & \textbf{49.30} & \textbf{34.24} & \textbf{64.62} \\
\midrule
Single-Winner Supervision & 20.99\% & 57.47 & 83.51 & 81.75 & \textbf{2203.5} & 83.80 & \textbf{52.47} & 56.20 & 54.40 & 68.54 \\
Tie-Aware Supervision & 20.99\% & \textbf{58.38} & \textbf{85.08} & \textbf{83.04} & 2189.5 & \textbf{84.86} & 52.40 & \textbf{59.20} & \textbf{57.04} & \textbf{69.77} \\
\midrule
Question Content Removed & 20.99\% & 58.42 & 84.59 & 82.65 & 2205.7 & 85.01 & 53.00 & 58.90 & 57.08 & 69.80 \\
Full Selector & 20.99\% & 58.38 & 85.08 & 83.04 & 2189.5 & 84.86 & 52.40 & 59.20 & 57.04 & 69.77 \\
Question and Spatial Inputs Removed & 20.99\% & 58.20 & 84.50 & 82.95 & 2195.8 & 84.77 & 53.07 & 58.10 & 57.20 & 69.65 \\
\bottomrule
\end{tabular}
}
\label{tab:ablations}
\end{table*}

Behavioral training improves mean pooling by 5.57 macro points (95\% CI $[5.00,6.20]$), with gains across all eight benchmarks. Tie-aware supervision improves over arbitrary single-winner targets by 1.24 points (95\% CI $[0.768,1.734]$), supporting set-valued supervision when multiple regions are equally optimal.

Question content does not provide a measurable gain: removing it yields 69.80 versus 69.77 for the full selector, with a paired difference of $-0.03$ points (95\% CI $[-0.238,0.180]$). Removing both question and explicit spatial inputs yields 69.65, although this also changes model capacity. Together with Section~\ref{sec:fidelity_analysis}, these results suggest that the limitation extends beyond question content alone to information accessible before language-model computation.


\section{Limitations}
\label{sec:limitations}

The current selector remains below strong whole-image resizing, and matched probing indicates that compression-sensitive signals are only weakly accessible before language-model computation. Moreover, preserving one native-resolution cell cannot recover information lost across the remaining compressed regions. Our token ratios measure visual tokens entering the language model, not end-to-end compute: Foveated Compression runs the full-resolution vision encoder once, whereas pixel-space downsampling also reduces vision-encoder cost. The comparison with downsampling is therefore favorable to the baseline in wall-clock terms, and we do not claim an end-to-end efficiency advantage. We do not report wall-clock or FLOPs measurements, and we do not benchmark against attention-based pruning methods such as FastV~\cite{fastv2024} and SparseVLM~\cite{sparsevlm2025} or crop-and-reprocess methods such as V*~\cite{vstar2024} and ZoomEye~\cite{zoomeye2025} under matched token and compute budgets. All spatial statistics, including the oracle gap, selector margins, and tie rates, are specific to the $3\times3$, $K=1$ action space; how they change with finer grids or larger $K$ remains untested. We evaluate a single frozen VLM backbone; broader backbones, finer action spaces, and adaptive budgets remain future work.


\section{Conclusion}
\label{sec:conclusion}

We introduced Foveated Compression, a fixed-budget mixed-resolution representation that combines global compression with native-resolution preservation from a single full-resolution encoding. Learned spatial allocation significantly improves over random and fixed selection, although strong whole-image resizing remains better overall. Within the same $3\times3$, $K=1$ action space, a budget-matched oracle reaches 82.73 versus 69.61 for the learned selector, revealing substantial selection headroom, while matched probing shows that compression-failure signals become substantially more accessible after language-model computation. These results suggest that exploiting spatial fidelity allocation will require selection signals that draw on language-model computation, together with stronger compressed representations.


\section{Acknowledgments}
\label{sec:acknowledgments}
This work was supported by a grant from the Institute of Information \& Communications Technology Planning \& Evaluation (IITP) grant funded by the Korean government (MSIT) (No. RS-2022-II220124, Development of Artificial Intelligence Technology for Self Improving Competency-Aware Learning Capabilities) and Electronics and Telecommunications Research Institute (ETRI) grant funded by the Korean government (26CB1200, Development and Application of Science-Specialized Multimodal Foundation Models).

\newpage

\section*{}
{
    \small
    \bibliographystyle{plainnat}
    \bibliography{main}
}

\newpage


\appendix
\setcounter{table}{0}
\setcounter{figure}{0}
\renewcommand{\thetable}{A\arabic{table}}
\renewcommand{\thefigure}{A\arabic{figure}}

\section{Implementation and Evaluation Details}
\label{app:implementation}

\subsection{Training Details}
\label{app:details}

We train the Foveated Merger and Foveated Selector sequentially while keeping the Qwen2.5-VL-7B-Instruct backbone frozen. Table~\ref{tab:training_details} summarizes the main optimization settings. The merger is trained for one epoch with 8-GPU distributed data parallelism, corresponding to 2,378 optimization steps, and uses gradient accumulation of four. We use AdamW with a learning rate of $1\times10^{-4}$, weight decay $1\times10^{-2}$, 200 warmup steps followed by a constant learning rate, and gradient clipping at 1.0. The training visual-token cap is 1,600 tokens, reduced to 600 for GQA. Teacher predictions are cached for 79,723 examples using the top-256 output probabilities. Main training and full-split evaluation runs were performed on 8× NVIDIA RTX A5000 24GB GPUs. Training the Foveated Merger took approximately 6.9 GPU-hours. Generating exhaustive K=1 selector labels required executing all nine deterministic K=1 actions per training example and took approximately 18.1 GPU-hours in total.

The selector is trained only after the merger has been frozen. It is trained for four epochs with a learning rate of $2\times10^{-4}$ and 100 warmup steps. Training and validation partitions are formed at the image-group level so that multiple questions associated with the same image do not cross splits. Question inputs are truncated to at most 48 tokens. All-nine-tie examples are excluded from selector training, while partial ties are retained through the set-valued objective in Equation~\eqref{eq:tied_winner_loss}.

\subsection{Evaluation Protocol}
\label{app:protocol}

All reported conditions, including the native reference, use the same generation protocol. We append the common instruction
\begin{quote}
\texttt{Answer the question using a single word or phrase.}
\end{quote}
to the benchmark question and decode greedily with sampling disabled, a single beam, and at most 16 generated tokens. Benchmark predictions are scored using each benchmark's official evaluation metric. MME is reported on its native 0--2800 scale in per-benchmark tables and divided by 28 only when computing the unweighted benchmark macro average.

The native reference uses the same 252-aligned source geometry as all Foveated Compression conditions. This differs from allowing each condition to independently choose its native preprocessing size and ensures that comparisons begin from the same visual input. Downsampling baselines use bicubic image resizing before visual encoding.

\subsection{Geometry and Token Accounting}
\label{app:geometry}

Image dimensions are chosen as multiples of 252 pixels. This alignment follows
\[
252
=
14
\times
2
\times
3
\times
3,
\]
corresponding to the 14-pixel vision patch size, Qwen's native $2\times2$ visual merge, the Foveated Merger's $3\times3$ reduction, and the coarse $3\times3$ selection grid. Consequently, compression blocks and selection-cell boundaries align exactly.

Table~\ref{tab:token_accounting} summarizes the visual-token budgets. Uniform Foveated Compression uses exactly $1/9=11.11\%$ of the native visual tokens. Restoring one of nine cells gives $17/81=20.99\%$ as derived in Equation~\eqref{eq:mixed_token_count}. The matched-budget downsampling baseline targets the same approximately 21\% budget but uses 19.82\% in practice because image dimensions are quantized to the model's spatial alignment. Similarly, halving both image dimensions would nominally produce a 25\% token budget, while the aligned $2\times$ Downsample condition uses 24.57\%.

\begin{table}[t]
\centering
\small
\caption{Main training settings for the Foveated Merger and Foveated Selector. Settings not shared across the two modules are marked with ``--''.}
\begin{tabular}{lcc}
\toprule
Setting & Merger & Selector \\
\midrule
Trainable parameters & 32.77M & 10.8M \\
Learning rate & $1\times10^{-4}$ & $2\times10^{-4}$ \\
Epochs & 1 & 4 \\
Warmup steps & 200 & 100 \\
Optimization steps & 2,378 & -- \\
Gradient accumulation & 4 & -- \\
Gradient clipping & 1.0 & -- \\
Question length & -- & 48 \\
\bottomrule
\end{tabular}
\label{tab:training_details}
\end{table}

\begin{table}[t]
\centering
\small
\caption{Visual-token budgets relative to the 252-aligned native input.}
\begin{tabular}{lc}
\toprule
Condition & Visual-token ratio \\
\midrule
Native & 100.00\% \\
Uniform Foveated Compression & 11.11\% \\
Iso-Token Downsample & 11.11\% \\
Random / Fixed / Learned $K{=}1$ & 20.99\% \\
Matched-Budget Downsample & 19.82\% \\
$2\times$ Downsample & 24.57\% \\
$K{=}1$ Region-Choice Oracle & 20.99\% \\
\bottomrule
\end{tabular}
\label{tab:token_accounting}
\end{table}

These ratios measure the number of visual tokens entering the language model. They should not be interpreted as end-to-end compute ratios: Foveated Compression executes the full-resolution vision encoder once, whereas pixel-space downsampling also reduces vision-encoder computation.

\section{Selector Supervision Statistics}
\label{app:selector_labels}

Selector supervision is generated only after freezing the trained Foveated Merger. For each training example, we execute all nine deterministic $K{=}1$ actions and record the set of cells attaining the maximum task score. Examples for which all nine cells tie are excluded because they contain no spatial preference; all other tied winners are retained.

Table~\ref{tab:selector_label_stats} summarizes the resulting supervision. Among 18,048 generated labels, 7,678 examples (42.54\%) are spatially informative for selector training. Informative examples contain 3.66 tied winners on average. Consequently, a uniformly random cell belongs to the winner set in 40.65\% of informative training examples, making random region selection a nontrivial baseline. The best single fixed cell chosen from training labels obtains 45.40\% winner membership, while the held-out learned selector reaches 55.34\%.

\begin{table}[t]
\centering
\small
\caption{Statistics of the intervention-derived selector supervision. Winner membership measures whether the selected cell belongs to the set of equally optimal $K{=}1$ actions.}
\begin{tabular}{lr}
\toprule
Statistic & Value \\
\midrule
Total selector labels & 18,048 \\
Spatially informative labels & 7,678 \\
Informative fraction & 42.54\% \\
Mean winner-set size & 3.66 \\
Random winner membership & 40.65\% \\
Best training-only fixed cell & 45.40\% \\
Held-out selector winner hit & 55.34\% \\
\bottomrule
\end{tabular}
\label{tab:selector_label_stats}
\end{table}

These statistics should not be confused with the 21.5\% spatially informative rate in Section~\ref{sec:hr_matter}. The latter is measured over the full 37,610-example evaluation set, whereas Table~\ref{tab:selector_label_stats} characterizes the separate data used to construct selector supervision.

\section{Additional Diagnostic and Statistical Results}
\label{app:additional_results}

\subsection{Resolution and Spatial-Choice Diagnostics}

Table~\ref{tab:diagnostic_summary} collects the aggregate diagnostics summarized in Section~\ref{sec:hr_matter}. Uniform Foveated Compression retains 86.1\% of predictions that are correct at native resolution, but sensitivity varies substantially by task: native-correct retention ranges from 38.8\% on ChartQA to 94.3\% on POPE.

Across exhaustive $K{=}1$ interventions, all nine regions obtain the same task score for 78.5\% of evaluation examples. The remaining 21.5\% are spatially informative, meaning that at least two region choices produce different outcomes. For informative examples with winner sets containing between two and eight cells, 41.9\% of winner pairs are horizontally or vertically adjacent. A uniformly random pair on a $3\times3$ grid is adjacent with probability
$12/\binom{9}{2}=1/3$.

\begin{table}[t]
\centering
\small
\caption{Aggregate resolution-sensitivity and spatial-choice diagnostics.}
\begin{tabular}{lr}
\toprule
Diagnostic & Value \\
\midrule
Evaluation examples & 37,610 \\
Overall native-correct retention & 86.1\% \\
Lowest retention (ChartQA) & 38.8\% \\
Highest retention (POPE) & 94.3\% \\
All-nine-tie examples & 78.5\% \\
Spatially informative examples & 21.5\% \\
Adjacent winner pairs & 41.9\% \\
Random-pair adjacency & 33.3\% \\
\bottomrule
\end{tabular}
\label{tab:diagnostic_summary}
\end{table}

The adjacency statistic is used as descriptive evidence of spatial structure rather than as a formal significance test.

\subsection{Paired Bootstrap Comparisons}

We quantify uncertainty using 2,000 paired bootstrap resamples. Evaluation rows are resampled within benchmark so that predictions from the compared methods remain paired, and the benchmark macro average is recomputed for each resample using the same MME normalization as in the main results. Table~\ref{tab:bootstrap_full} reports the principal comparisons.

\begin{table*}[t]
\centering
\small
\caption{Paired bootstrap comparisons. Positive values favor the first method named in each comparison. The oracle is a non-deployable test-time upper bound.}
\begin{tabular}{lrr}
\toprule
Comparison & $\Delta$ Macro & 95\% CI \\
\midrule
Uniform Foveated Compression $-$ Iso-Token Downsample
& $-0.24$ & $[-0.586,\;0.521]$ \\
\midrule
Foveated Selector $-$ Native
& $-9.238$ & $[-9.867,\;-8.625]$ \\
Foveated Selector $-$ Random $K{=}1$
& $+1.622$ & $[1.085,\;2.161]$ \\
Foveated Selector $-$ Fixed $K{=}1$
& $+1.131$ & $[0.610,\;1.632]$ \\
Foveated Selector $-$ Matched-Budget Downsample
& $-1.378$ & $[-2.062,\;-0.707]$ \\
$2\times$ Downsample $-$ Foveated Selector
& $+2.802$ & $[2.164,\;3.438]$ \\
\midrule
Oracle $-$ Native
& $+3.873$ & $[3.388,\;4.366]$ \\
Oracle $-$ Foveated Selector
& $+13.124$ & $[12.566,\;13.691]$ \\
Oracle $-$ Random $K{=}1$
& $+14.753$ & $[14.173,\;15.350]$ \\
Oracle $-$ Fixed $K{=}1$
& $+14.242$ & $[13.688,\;14.814]$ \\
Oracle $-$ Matched-Budget Downsample
& $+11.747$ & $[11.143,\;12.350]$ \\
\midrule
Behaviorally Trained Merger $-$ Mean Pooling
& $+5.57$ & $[5.00,\;6.20]$ \\
Tie-Aware Supervision $-$ Single-Winner Supervision
& $+1.236$ & $[0.768,\;1.734]$ \\
Full Selector $-$ Question Content Removed
& $-0.031$ & $[-0.238,\;0.180]$ \\
\bottomrule
\end{tabular}
\label{tab:bootstrap_full}
\end{table*}

Uniform Foveated Compression and Iso-Token Downsample show no significant paired difference at the 11.11\% budget; we do    not claim equivalence. At the $K{=}1$ operating point, the learned selector significantly improves over random and fixed allocation but remains below both matched-budget and $2\times$ whole-image resizing. The large oracle gap therefore reflects unrealized region-choice headroom rather than an advantage achieved by the current selector.

\section{Matched Compression-Failure Analysis}
\label{app:matched_auc}

Section~\ref{sec:fidelity_analysis} compares the accessibility of compression-sensitive signals before and after language-model computation. To make this comparison directly interpretable, both feature families are evaluated on the same examples and the same binary target.

The matched set contains 3,416 examples, comprising 488 examples from each of seven benchmarks. We define
\begin{equation}
y_{\mathrm{rescue}}
=
\mathbb{1}
\left[
\text{native correct}
\land
\text{uniform compression incorrect}
\right].
\label{eq:app_native_rescue}
\end{equation}
There are 178 positive examples, corresponding to 5.21\% of the matched set. The target therefore measures whether aggressive compression changes a correct native prediction into an incorrect one. It does \emph{not} identify which of the nine $K{=}1$ regions should be selected.

For the primary raw comparison, the prefill-free signal is the image-level mean of the selector-side keep logits, while the decoder-side signal is the first-token entropy from generation under uniform compression. Both are evaluated on the same rows. We additionally train same-fold five-fold out-of-fold probes for the prefill-free and decoder-side feature families. Confidence intervals and paired differences are estimated with 2,000 benchmark-stratified paired bootstrap resamples.

\begin{table}[t]
\centering
\small
\caption{Matched AUC comparison for predicting compression-induced answer failure.}
\begin{tabular}{lcc}
\toprule
Signal & AUC & 95\% CI \\
\midrule
Prefill-free raw signal
& 0.5239 & $[0.4796,\;0.5700]$ \\
First-token entropy
& 0.7651 & $[0.7347,\;0.7937]$ \\
\midrule
Prefill-free OOF probe
& 0.5476 & $[0.5012,\;0.5929]$ \\
Decoder-side OOF probe
& 0.8072 & $[0.7738,\;0.8391]$ \\
\bottomrule
\end{tabular}
\label{tab:matched_auc}
\end{table}

The raw decoder signal exceeds the raw prefill-free signal by 0.2411 AUC (95\% CI $[0.1845,0.3015]$). The out-of-fold comparison gives a similar advantage of 0.2596 (95\% CI $[0.2034,0.3132]$). Across individual decoder statistics, negative minimum-token log-probability achieves the highest observed AUC, 0.8105, whereas raw prefill-free selector statistics range from 0.4782 to 0.5239.

These results should not be interpreted as showing that decoder uncertainty directly solves region selection. First, the binary target asks \emph{whether} native fidelity is needed rather than \emph{where} it should be allocated. Second, decoder uncertainty becomes available only after language-model generation has begun and therefore cannot be used directly by the single-pass prefill-free selector studied in this work. The matched comparison instead shows that information associated with compression-induced failure is substantially more accessible after language-model computation.


\begin{figure*}[t]
    \centering

    \begin{subfigure}[t]{0.32\textwidth}
        \centering
        \includegraphics[width=\linewidth]{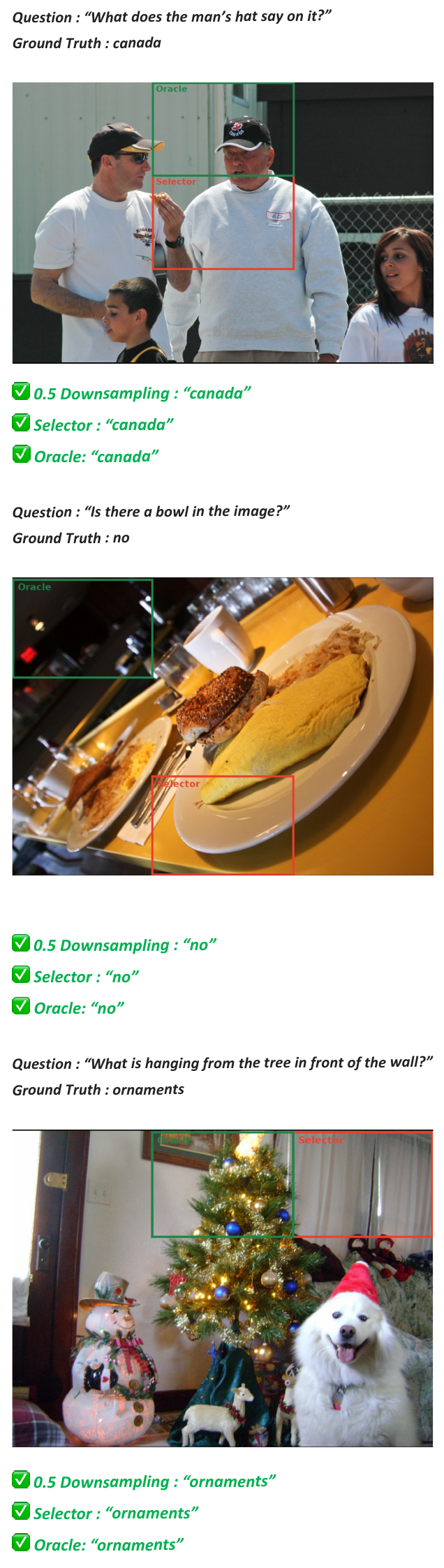}
        \caption{Compression-robust}
        \label{fig:qualitative_a}
    \end{subfigure}
    \hfill
    \begin{subfigure}[t]{0.32\textwidth}
        \centering
        \includegraphics[width=\linewidth]{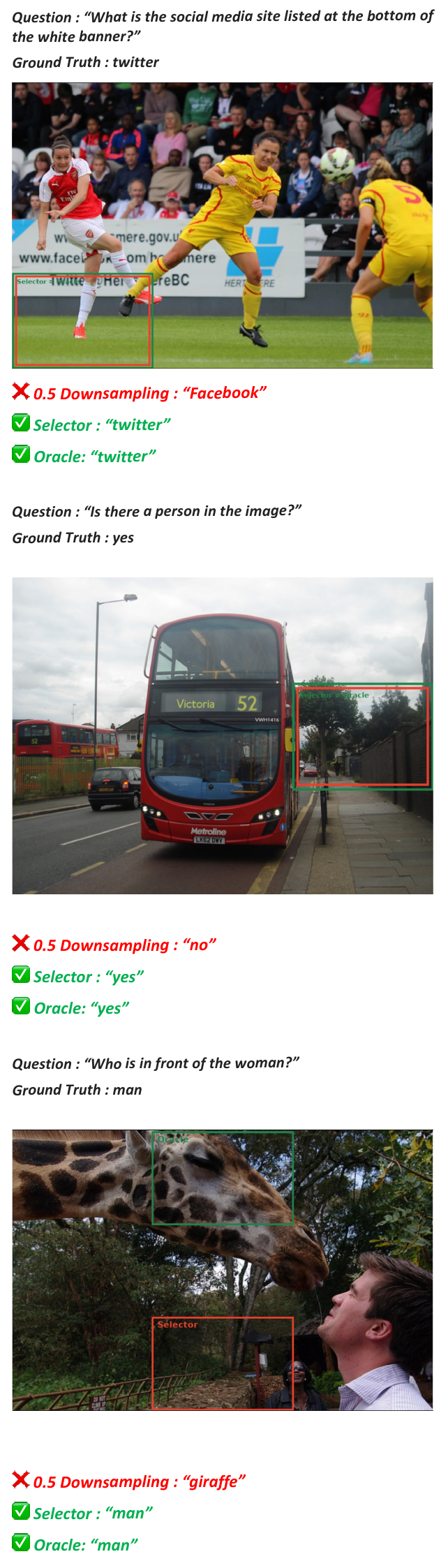}
        \caption{Selection-rescued}
        \label{fig:qualitative_b}
    \end{subfigure}
    \hfill
    \begin{subfigure}[t]{0.32\textwidth}
        \centering
        \includegraphics[width=\linewidth]{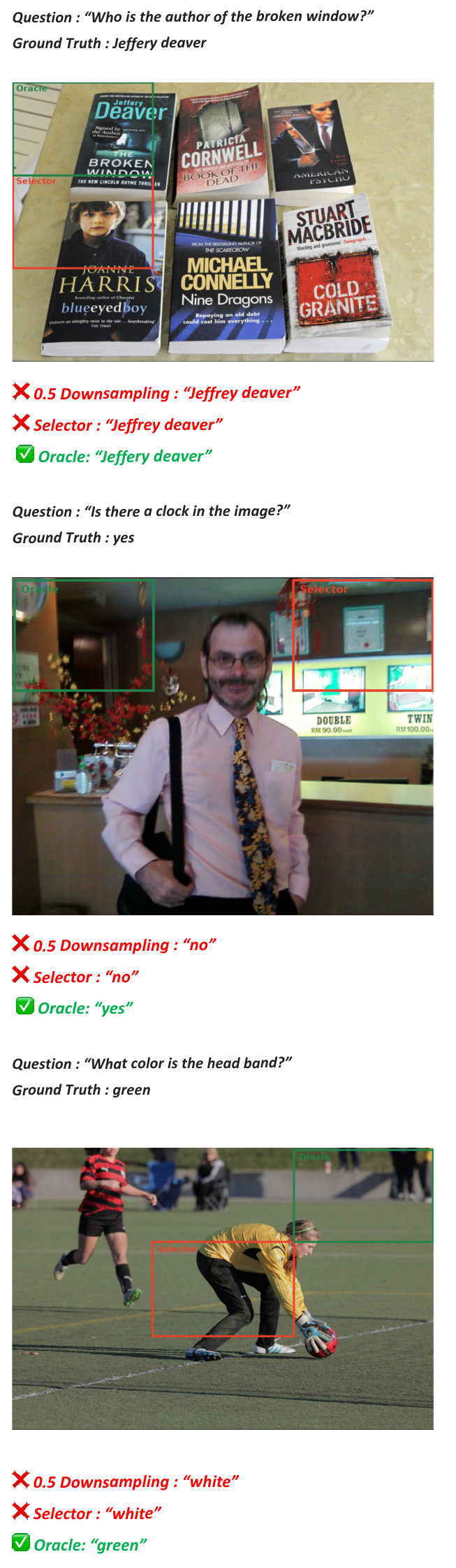}
        \caption{Selection-miss}
        \label{fig:qualitative_c}
    \end{subfigure}

    \caption{
    Additional qualitative examples across three outcome regimes.
    Each column contains examples of scene-text reading, object existence,
    and compositional reasoning.
    \textbf{(a) Compression-robust:} downsampling, the learned selector,
    and the $K{=}1$ oracle all succeed.
    \textbf{(b) Selection-rescued:} downsampling fails, while the learned
    selector preserves a region that recovers the correct answer.
    \textbf{(c) Selection-miss:} both downsampling and the learned selector
    fail, but another $K{=}1$ oracle action succeeds.
    Oracle regions are shown only as a diagnostic upper bound and are not
    available to the deployable selector.
    }
    \label{fig:qualitative}
\end{figure*}

\section{Additional Qualitative Examples}
\label{app:qualitative}

We provide additional qualitative examples to complement the aggregate analysis in Section~\ref{sec:fidelity_analysis}. Figure~\ref{fig:qualitative} presents nine examples organized into three outcome regimes: \emph{compression-robust} (Figure~\ref{fig:qualitative_a}), \emph{selection-rescued} (Figure~\ref{fig:qualitative_b}), and \emph{selection-miss} (Figure~\ref{fig:qualitative_c}). Each regime contains one example from each of three representative visual reasoning categories---scene-text reading, object existence, and compositional reasoning---illustrating that these outcome patterns are not limited to a single type of visual evidence.

For each example, we show the question and ground-truth answer together with predictions from whole-image downsampling, the learned Foveated Selector, and the best $K{=}1$ oracle action. We additionally visualize the regions selected by the learned selector and the oracle when they differ. The oracle is included only as a diagnostic upper bound that identifies whether a successful action exists within the same fixed-budget $K{=}1$ action space; it should not be interpreted as a deployable prediction method.

In the \emph{compression-robust} cases (Figure~\ref{fig:qualitative_a}), aggressive resolution reduction does not change the task outcome: whole-image downsampling, the learned selector, and the oracle all produce successful predictions. These examples are consistent with our aggregate observation that many inputs are insensitive to the precise allocation of native-resolution fidelity. In such cases, allocating additional native-resolution tokens to a particular region provides little benefit because the evidence required for the prediction remains sufficiently accessible under global resolution reduction.

In the \emph{selection-rescued} cases (Figure~\ref{fig:qualitative_b}), whole-image downsampling fails, whereas the learned selector identifies a region whose native-resolution preservation enables the mixed-resolution representation to recover the correct answer. The examples span scene-text reading, object existence, and compositional reasoning, showing that selective preservation can recover information lost by uniform resizing across qualitatively different forms of visual evidence. These cases illustrate the benefit that learned spatial fidelity allocation can provide under the same fixed-budget representation when task-relevant information is localized to a region that benefits from higher visual fidelity.

Finally, the \emph{selection-miss} cases (Figure~\ref{fig:qualitative_c}) show examples for which both whole-image downsampling and the learned selector fail, while another $K{=}1$ action selected by the oracle produces the correct answer. Because the learned and oracle conditions use the same mixed-resolution representation and differ only in which spatial cell is preserved, these examples isolate region selection as the source of failure. They qualitatively illustrate the substantial selector--oracle gap reported in Table~\ref{tab:foveated_main_results}: the fixed $K{=}1$ action space can contain a successful allocation even when the learned selector fails to identify it.

Together, these examples expose three complementary regimes of fixed-budget fidelity allocation. Some examples remain robust to aggressive compression, some benefit from the region selected by the learned model, and others could be recovered within the same representation if region selection were more accurate. This qualitative progression is consistent with our quantitative findings that spatial fidelity allocation has meaningful potential, while improved region selection remains necessary to exploit the available headroom.


\clearpage

\end{document}